\documentclass[conference]{IEEEtran}

\usepackage{cite}
\usepackage{graphicx}
\usepackage{booktabs}
\usepackage{multirow}
\usepackage[colorlinks,urlcolor=blue,citecolor=red,bookmarks=false]{hyperref}
\usepackage{flushend}
\usepackage{xpunctuate}

\providecommand{\eg}{\emph{e.g}\xperiod}
\providecommand{\ie}{\emph{i.e}\xperiod}
\providecommand{\etc}{\emph{etc}\xperiod}
\providecommand{\etal}{\emph{et al}\xperiod}

\begin{document}

\title{Color Mismatches in Stereoscopic Video: Real-World Dataset and Deep Correction Method}

\author{
    \IEEEauthorblockN{Egor Chistov \quad Nikita Alutis \quad Dmitriy Vatolin}
    \IEEEauthorblockA{\textit{Lomonosov Moscow State University}}
    \IEEEauthorblockA{\texttt{\{egor.chistov, nikita.alutis, dmitriy\}@graphics.cs.msu.ru}}
}

\maketitle

\begin{abstract}

Stereoscopic videos can contain color mismatches between the left and right views due to minor variations in camera settings, lenses, and even object reflections captured from different positions. The presence of color mismatches can lead to viewer discomfort and headaches. This problem can be solved by transferring color between stereoscopic views, but traditional methods often lack quality, while neural-network-based methods can easily overfit on artificial data. The scarcity of stereoscopic videos with real-world color mismatches hinders the evaluation of different methods' performance. Therefore, we filmed a video dataset, which includes both distorted frames with color mismatches and ground-truth data, using a beam-splitter. Our second contribution is a deep multiscale neural network that solves the color-mismatch-correction task by leveraging stereo correspondences. The experimental results demonstrate the effectiveness of the proposed method on a conventional dataset, but there remains room for improvement on challenging real-world data. 

\end{abstract}

\begin{IEEEkeywords}
stereoscopic 3D, real-world datasets, deep learning, color transfer
\end{IEEEkeywords}

\section{Introduction}

\begin{figure}[t]
    \centering
    \includegraphics[width=\linewidth]{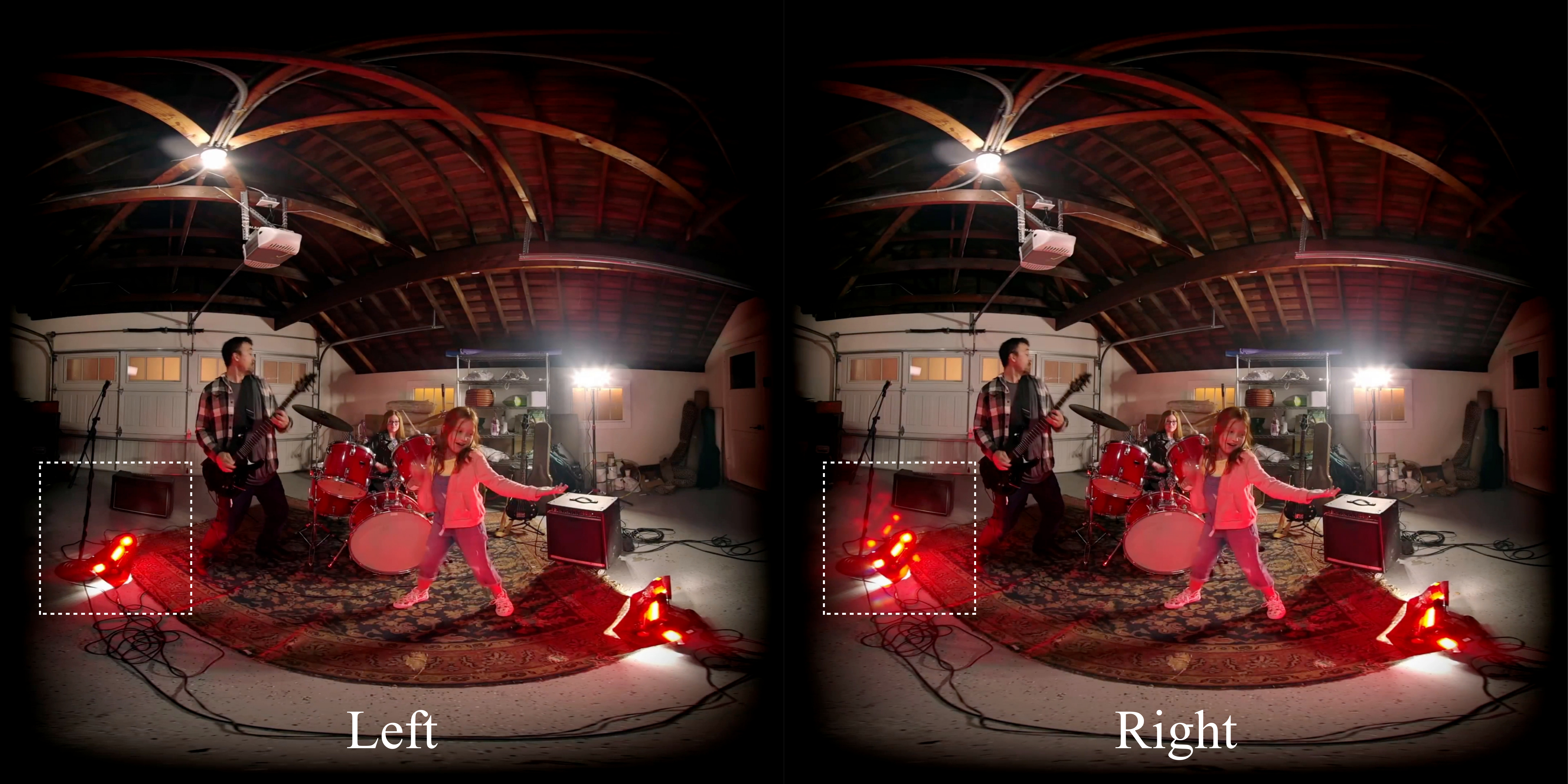}
    \caption{Frame \#1,200 from video “VR180 Cameras with Daydream,” taken by Google (\url{https://www.youtube.com/watch?v=TH_MMXinRsA}), contains color mismatches.}
    \label{fig:example}
\end{figure}

The left and right views of a stereoscopic image (stereopair) can have color mismatches for various reasons, \eg glares and polarized light; Figure~\ref{fig:example} shows an example of a glare. Color mismatches can degrade the viewing experience and cause discomfort, including eye strain, blurred vision, and a burning sensation in the eyes~\cite{antsiferova2017influence, zeri2015visual, cho2014visual, lambooij2011susceptibility}. Thus, color correction algorithms~\cite{lavrushkin2018local, niu2019visually, croci2021deep} have been proposed to eliminate color discrepancies in stereoscopic images and videos.

Besides algorithms, datasets were created for color-mismatch correction~\cite{hwang2014color, niu2016image, lavrushkin2018local, croci2021deep}. The limited availability of these datasets hinders the comparison of new color transfer methods. Also, they are either small or lack real-world mismatches. A sufficiently large dataset and sampling of color-mismatch types are crucial to comparing color-mismatch-correction methods. Therefore, we created a new dataset using a beam splitter and three cameras. We employed a beam splitter to generate distorted and ground-truth versions of the left view, while capturing the right view using a third camera. To ensure accurate ground-truth data, we employed a two-stage processing pipeline. Firstly, we aligned the cameras manually, using a photo from one camera and a video stream from another. Secondly, we used a homography transformation for fine-grain alignment and rectification. We also performed a temporal alignment using the audio streams captured by the three cameras. Our dataset contains 14 scenes, each seven frames long, with various objects that produced color mismatches. We used this dataset to compare different correction methods in the wild.

Despite the rapid development of deep learning, we think that the potential of neural networks is not yet fully utilized for the color-correction task. We developed a new deep method that transfers color by leveraging estimated correspondences. Given two input images, we computed an optical flow operator and a binary confidence map. The network then extracts features from input images, matches target and reference features using estimated optical flow, and passes these features along with a confidence map to a U-Net~\cite{ronneberger2015u} decoder. This is different from the Croci~\etal work~\cite{croci2021deep} that used an end-to-end neural network to solve both color-correction and correspondence-estimation tasks. We will show that our method demonstrates better results.

Existing color-transfer methods considered different color-distortion models. Evaluation on known models is usually used to find the upper bound of the methods' quality. But how do the methods generalize to unseen distortions? To explore this, we not only used the conventional artificial dataset, but also used our real-world dataset for an evaluation of eight methods. For trainable methods, we considered two training procedures---with deterministic and probabilistic distortion models---to find that that allows for better generalization. Our experiments showed that the probabilistic approach improved generalization to real-world videos. On the artificial dataset, our method was ranked the best by all quality-assessment methods with both training procedures. However, on the real-world data, all methods, which consider different non-trivial distortion models, performed worse than simple global methods. We demonstrated that this discrepancy was due to domain shift between the distortion model used in methods' development and a real-world distortion model.

Our contributions are summarized as follows: (1) We filmed a real-world dataset of stereoscopic videos. (2) We developed a new method for color correction. (3) We compared eight color-correction methods on two datasets. (4) We further demonstrate the discrepancies between results on these datasets. The code and datasets are at \url{https://github.com/egorchistov/color-transfer}.

\section{Related Work}

The color-transfer problem involves finding an image $I$ that has the same structure (\ie contours and textures) with the target image $T$ and the same color (\ie brightness, contrast, \etc) with the reference image $R$. This problem is solvable globally or locally. The global methods~\cite{reinhard2001color, xiao2006color, pitie2007linear} estimate a single color transformation for all pixels. In contrast, local methods~\cite{pitie2007automated, lavrushkin2018local, grogan2019l2, croci2021deep} estimate multiple color transformations, one for each pixel or image region. Local methods can be further divided into two categories: those based on the correspondence estimation~\cite{lavrushkin2018local, grogan2019l2, croci2021deep} and those that make no assumptions about the common part between images~\cite{pitie2007automated}. The pioneering work of Reinhard \etal~\cite{reinhard2001color} used the following form of color transfer:
\begin{equation}
    I = (T - \mu_{T}) \Sigma_R \Sigma_{T}^{-1} + \mu_R,
\end{equation}
where $\mu$ is the mean of the colors in the palette and $\Sigma$ is the diagonal covariance matrix for the colors in the palette. Hereinafter, we will employ the broadcast notation, \eg the operation of subtracting the number $\mu_{T}$ from the matrix $T$ is defined as the subtraction of this number from each element of the matrix. Xiao and Ma~\cite{xiao2006color} computed an arbitrary covariance matrix $\Sigma$ instead of the diagonal one. A singular value decomposition of the covariance matrix $\Sigma = USV$ was performed, resulting in unitary rotation matrices $U$ and $V$ and a diagonal scaling matrix $S$. The color transfer was then performed using the following formula:
\begin{equation}
    I = (T - \mu_{T}) U_{T} \sqrt{S_R S_{T}^{-1}} U_R^{-1} + \mu_R.
\end{equation}
There are the other variants of the covariance matrix decomposition for this problem~\cite{pitie2007linear}. More complex approaches include the local color transfer method of Pitié \etal~\cite{pitie2007automated}, that iteratively projects the target image $T$ and the reference image $R$ onto random one-dimensional axes and performs a probability-density transfer along those axes. Grain-noise artifacts arising from such a transfer were reduced by minimizing the difference of gradients between the resulting image $I$ and the target image $T$. Recent correspondence-based methods can utilize different types of matching algorithms. Block-matching-based algorithms~\cite{lavrushkin2018local} modify the matching-cost function by subtracting the block average before blocks are compared. This modification is necessary to successfully deal with color mismatches. In contrast, neural network-based matching algorithms~\cite{wang2020parallax} do not need such modification due to the robustness of neural networks to color changes in input images. Once the correspondences between the target image and the reference image have been established, a number of color transfer variants can be considered. Lavrushkin \etal~\cite{lavrushkin2018local} applied a guided filter~\cite{he2012guided} with a variable-length kernel, using the matched reference image as a guidance. Grogan and Dahyot~\cite{grogan2019l2} minimized a divergence between two probability-density functions of matched images. Croci \etal~\cite{croci2021deep} employed a neural network comprising six residual blocks to transfer color. We will compare these approaches in Section~\ref{sec:experiments}.

Few datasets were created for color transfer and color correction~\cite{hwang2014color, niu2016image, lavrushkin2018local, croci2021deep}. The existing datasets have several limitations: they are not publicly available, making it difficult to compare new methods, and they are either small or lack real-world color mismatches. The real-world dataset was created by Hwang \etal~\cite{hwang2014color}. They chose 15 image pairs captured by one camera under different illumination conditions, camera settings, and color touch-up styles. Reference images were aligned to match the target images. Artificial datasets are more widespread. For example, Niu \etal~\cite{niu2016image} employed 2D videos with parallel camera motion to extract pseudo-stereopairs without color mismatches; they gathered 18 source stereopairs. Using Photoshop CS6, they applied six different color operators—each with three severity levels—to one view of the stereopair. Lavrushkin \etal~\cite{lavrushkin2018local} proposed gathering frames from stereoscopic movies produced only via 3D rendering to obtain undistorted stereopairs. They employed 1,000 FullHD frames. After adding color noise to one view of the stereopair, they smoothed the result by applying a domain transform filter~\cite{gastal2011domain}. The work of Croci \etal~\cite{croci2021deep} used 1,035 stereopairs without color distortions from three different datasets. They applied six color operators—each with six severity levels—to one view of the stereopair using Photoshop 2021.

\section{Proposed Dataset}\label{sec:dataset}

We propose a new real-world dataset of stereoscopic videos for evaluating color-mismatch-correction methods. We collected it using a beam splitter and three cameras that simultaneously capture three views of a scene: a distorted left view, the ground-truth left view, and a right view. The beam splitter distortions we captured are similar to those commonly found in stereoscopic movies.

\begin{figure}[tb]
    \centering
    \includegraphics[width=\linewidth]{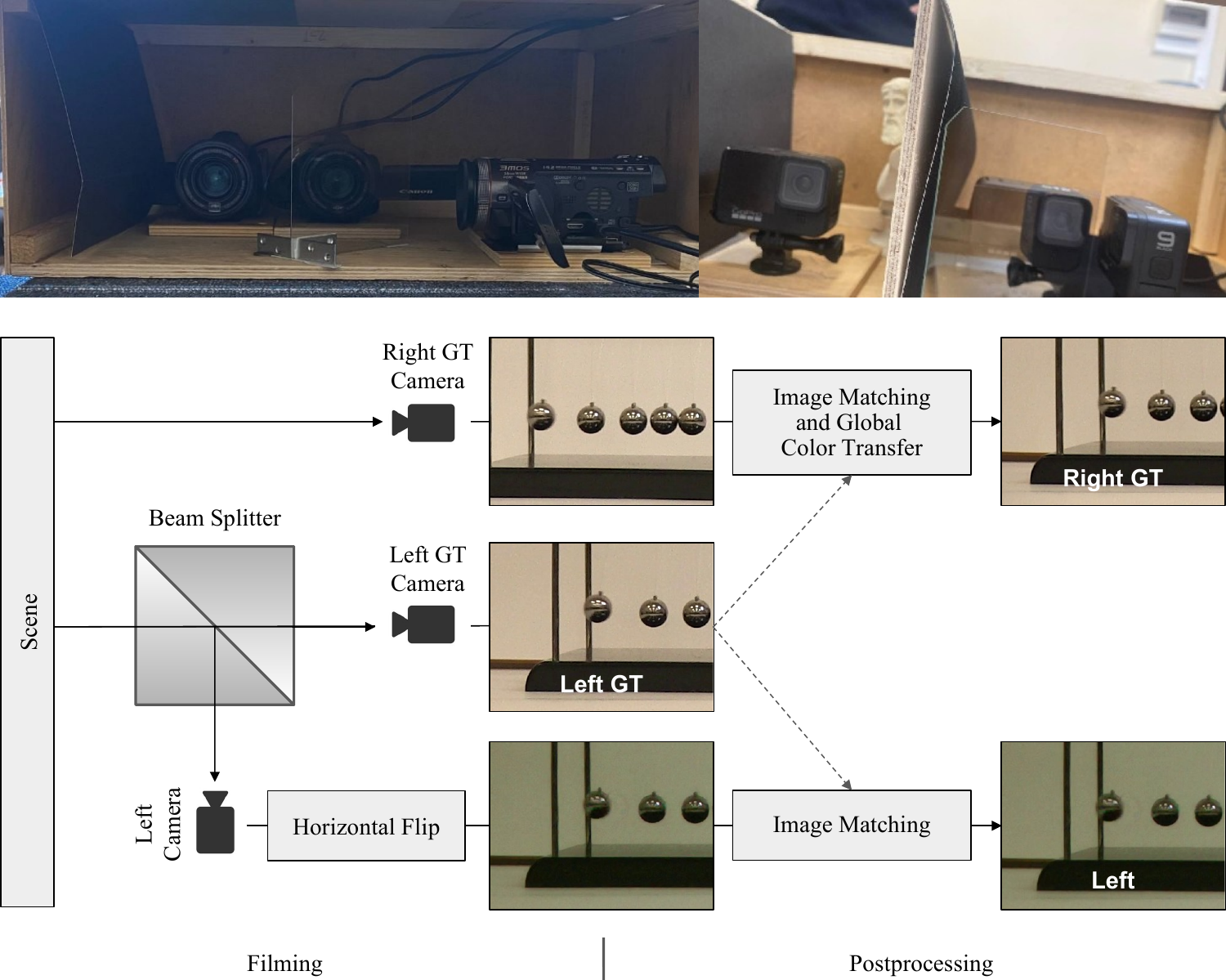}
    \caption{Two our setups in the top row and dataset filming and postprocessing pipeline in the bottom row. A beam splitter divides incoming light, which is then captured by the left camera and left ground-truth camera. The right camera captures the right ground-truth view. The final dataset frames have undergone spatial and temporal alignment.}
    \label{fig:pipeline}
\end{figure}

A beam splitter is an optical device that divides light into two beams. Our device, which split the beams unevenly, inherently introduced real-world mismatches between stereopair views, resembling the imperfections found in actual stereoscopic systems. We used a beam splitter to create a distorted ground-truth data pair by setting a zero stereobase between the left camera and the left ground-truth camera. A third camera captured the right view. Figure~\ref{fig:pipeline} shows our setup. We disabled optical stabilization and manually assigned all available camera settings, such as ISO, shutter speed, and color temperature. The cameras all had identical settings, so we obtained only beam-splitter distortions.

Filming the left distorted view and left ground-truth view without parallax is crucial to achieving precise ground-truth data. Parallax is a difference in the apparent position of an object when viewed along two lines of sight. Postprocessing can precisely correct affine mismatches between views—namely scale, rotation angle, and translation—but it cannot correct parallax without using complex optical-flow, occlusion-estimation, and inpainting algorithms, which can reduce ground-truth-data quality. We encountered a slight parallax due to inaccurate mounting of the cameras in our setup. To minimize it, we first visually aligned the lenses of the left distorted and left ground-truth cameras to obtain a visual zero stereobase. Then, using a photo from one camera and a video stream from another, we visualized the squared error between the views. This visualization allowed us to maximize the overlapping region by manually moving one camera. We achieved zero parallax through this method, but the manual alignment was time consuming.

Our research employed the postprocessing pipeline in Figure~\ref{fig:pipeline}. Given the three videos from our cameras, we matched the horizontally flipped left distorted view to the left ground-truth view and rectified the stereopair using a homography transformation. To estimate the transformation parameters, we employed MAGSAC++~\cite{barath2020magsac++}. We selected SIFT~\cite{lowe2004distinctive} to match the left views, as well as the LoFTR~\cite{sun2021loftr} for rectification because LoFTR can better handle large parallax. The technique of Pitié and Kokaram~\cite{pitie2007linear} perfectly corrected global color difference between ground-truth views. Without it, our approach would have been unable to obtain accurate ground-truth data. We performed temporal alignment using audio streams captured by the cameras: superimposing one stream on the other in the video editor allowed us to find the offset in frames.

\begin{figure}[tb]
    \centering
    \includegraphics[width=\linewidth]{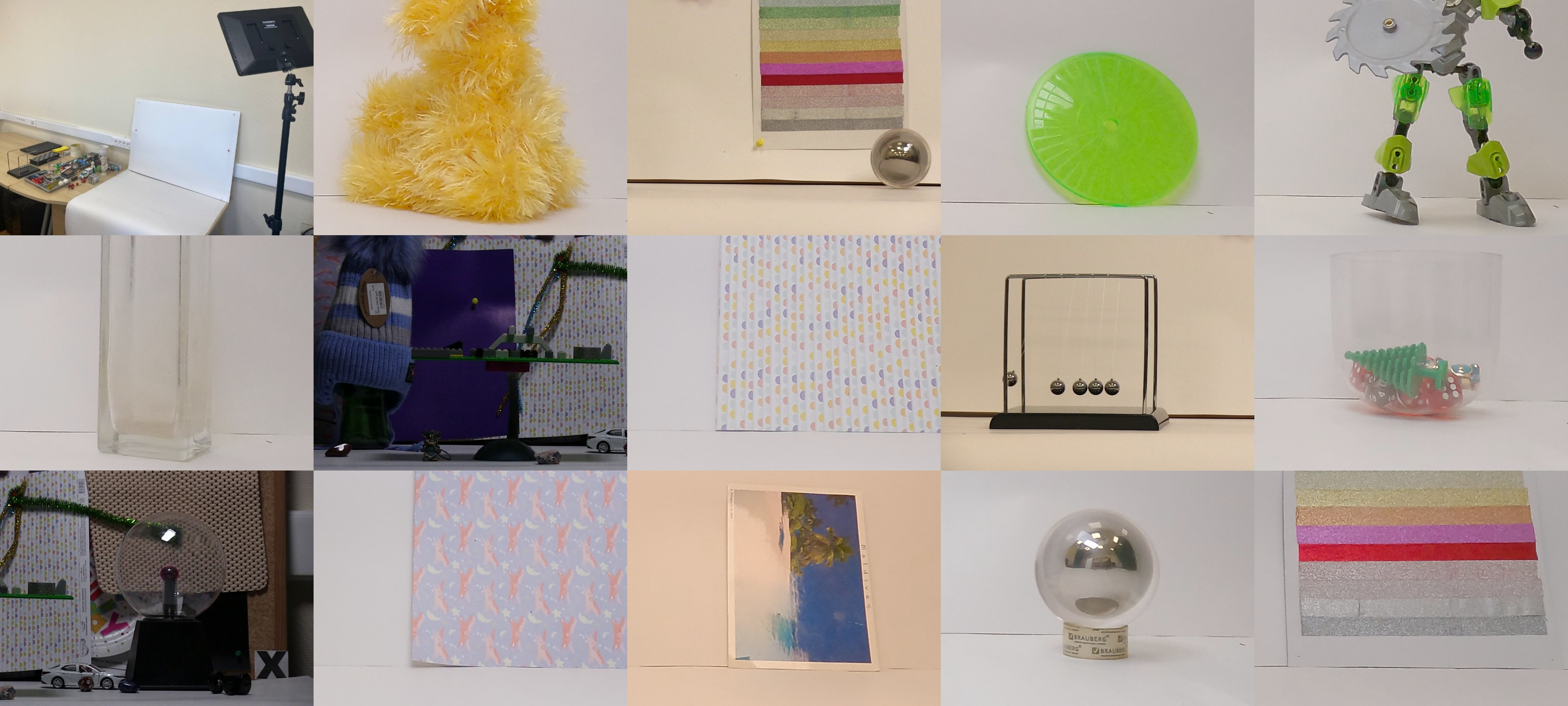}
    \caption{Our filming setup in the top left corner and all 14 scenes from our real-world dataset. It contains various objects that produce color mismatches. The dataset is publicly available at the project page.}
    \label{fig:scenes}
\end{figure}

Our research involved scenes in which an object rests on a table with a white background, 1.5 meters from the camera (Figure~\ref{fig:scenes}). We filmed various objects including transparent glass, color patterns, and moving objects, choosing those that produced color mismatches. To ensure diversity, we filmed scenes under different lighting conditions, moving at different speeds, and we also used different cameras to ensure distortion variety. The lighting changed dynamically in one scene, and different scenes were shot with different set lighting. We recorded most scenes in 4K resolution using three GoPro Hero 9 cameras. For the rest we recorded FullHD on two Legria HF G-10 cameras, one for the left ground-truth view and one for the right view, as well as a Panasonic HDC-SD800 for the left distorted view to add different camera distortions. The result was 14 scenes, each seven frames long. We then cropped the region of interest to 960x720 pixels.

\section{Proposed Method}

Color-mismatch correction involves transferring color from one view of a stereopair to another in areas where colors incorrectly differ. Without limiting the task’s generality, we constructed a deep neural network that uses the left view as a target and a right view as a reference. The resulting view has a structure consistent with that of the left view and the colors consistent with those of the right view.

From the two input images $T$ and $R$, we computed the optical flow operator $M$ and the binary confidence map $V$. Current state-of-the-art neural network-based optical flow estimation methods, namely RAFT~\cite{teed2020raft} and GMFlow~\cite{xu2022gmflow}, can successfully handle images with color mismatches. In this work, we chose GMFlow because it was trained on a larger dataset collection. There is also a reason, why we choose the optical flow algorithm for a stereo-correspondences estimation, even there is such methods as RAFT-Stereo~\cite{lipson2021raft} and GMStereo~\cite{xu2023unifying}. These methods heavily rely on an epipolar line being horizontal, \ie they scan for matching pixels only in the horizontal direction, which is often not the case for 3D, because of rectification algorithm failures~\cite{wang2023practical}. There are even curved epipolar lines, \eg in VR180. The result of correspondence estimation on non-rectified images is unsatisfactory, as shown in Figure 4~(b) of~\cite{zhao2023dive}.

If the target image $T$ was entirely contained within the reference image $R$ (\eg if the target image was a fragment of the reference image), the ideal optical flow operator $M^*$ would be able to correctly map all regions, and the color transfer would be expressed as follows: 
\begin{equation}
    I = M^*R.
\end{equation}

However, stereoscopic views overlap incompletely. For non-matched pixels $V = 0$, it is necessary to compute the colors based on neighboring matched pixels. This idea can be formulated as follows: 
\begin{equation}
    I = M^*R \odot V^* + f(T, M^*R, V^*) \odot (1 - V^*),
\end{equation}
where $\odot$ is the Hadamard (or element-wise) matrix product and $f$ is a function that computes the color based on the colors of neighboring matched pixels. This formulation requires a precise optical flow operator $M^*$, otherwise the matching errors will negatively impact the resulting image.

Since such quality is unattainable in practice, it is reasonable to introduce a new function $g$ to correct errors for matched pixels: 
\begin{equation}
    I = g(T, MR, V) \odot V + f(T, MR, V) \odot (1 - V).
\end{equation}

This formula is susceptible to the introduction of stitching artifacts on the edges of confidence regions. This can be overcome by introducing a new function $h$, which combines both $f$ and $g$:
\begin{equation}
    I = h(T, MR, V).
\end{equation}

We can approximate the function $h$ using a neural network. The choice of network architecture is limited only by the input and output shapes. However, the convolutional multi-scale architectures are preferable because they have the locality inductive bias, which means that the network will use the neighboring pixels to estimate the color of the given pixel. Standard segmentation architectures, such as U-Net~\cite{ronneberger2015u} and DeepLab~V3~\cite{chen2017rethinking}, are well suited for the task. We used the EfficientNet-B2~\cite{tan2019efficientnet} encoder to extract the features from the input images $T$ and $R$, then matched the target and reference features using the estimated optical flow operator $M$, and passed the matched features and the confidence map $V$ to the U-Net decoder to produce the resulting image $I$. We used the network of depth four in our experiments.

The network was trained using a linear combination of the mean squared error (MSE) and structural similarity (SSIM)~\cite{wang2004image} error: 
\begin{equation}
    \textit{MSE}(h(T, MR, V), GT) - \gamma \textit{SSIM}(h(T, MR, V), GT),
\end{equation}
where $GT$ denotes the ground-truth image and the parameter $\gamma$ was chosen so that both summands produce the same gradient in modulus during training. A value of $\gamma = 0.1$ was used.

\begin{table*}[t]
    \caption{Comparison of eight color-mismatch-correction methods on two datasets. The best result appears in \textbf{bold}. (C) stands for methods that were trained using Croci's training procedure and generalized worse to the real-world data.}
    \label{tab:comparison}
    \centering
    \begin{tabular}{l c c c c c c c c c}
\toprule
\multirow{2}[2]{*}{Method} & \multirow{2}[2]{*}{Type} & \multicolumn{4}{c}{Artificial Dataset} & \multicolumn{4}{c}{Real-World Dataset} \\
\cmidrule(lr){3-6}\cmidrule(lr){7-10}
 & & PSNR & SSIM & FSIM & iCID & PSNR & SSIM & FSIM & iCID \\
\midrule
Reinhard \etal~\cite{reinhard2001color} & Global & 34.03 & 0.960 & 0.984 & 0.124 & 32.28 & 0.931 & 0.958 & 0.171 \\
Xiao and Ma~\cite{xiao2006color} & Global & 33.11 & 0.951 & 0.982 & 0.161 & 31.12 & 0.919 & 0.952 & 0.217 \\
Piti{\'e} and Kokaram~\cite{pitie2007linear} & Global & 34.11 & 0.958 & 0.985 & 0.124 & \textbf{32.60} & \textbf{0.937} & \textbf{0.959} & \textbf{0.166} \\
Piti{\'e} \etal~\cite{pitie2007automated} & Local & 31.02 & 0.949 & 0.974 & 0.168 & 31.17 & 0.931 & 0.949 & 0.173 \\
Lavrushkin \etal~\cite{lavrushkin2018local} & Local & 31.84 & 0.956 & 0.980 & 0.146 & 28.83 & 0.927 & 0.949 & 0.189 \\
Grogan and Dahyot~\cite{grogan2019l2} & Local & 32.84 & 0.960 & 0.983 & 0.130 & 30.14 & 0.926 & 0.953 & 0.188 \\
Croci \etal~\cite{croci2021deep} & NN-based & 33.02 & 0.979 & 0.984 &  0.084 & 26.86 & 0.885 & 0.913 & 0.282 \\
Ours & NN-based & \textbf{35.26} & \textbf{0.988} & \textbf{0.992} & \textbf{0.073} & 26.00 & 0.913 & 0.921 & 0.272 \\
\midrule
Croci \etal~\cite{croci2021deep} (C) & NN-based & 37.72 & 0.989 & 0.992 & 0.036 & 22.26 & 0.875 & 0.906 & 0.345 \\
Ours (C) & NN-based & 39.68 & 0.995 & 0.997 & 0.029 & 23.21 & 0.903 & 0.916 & 0.333 \\
\bottomrule
\end{tabular}
\end{table*}

\begin{figure}[tb]
    \centering
    \includegraphics[width=\linewidth]{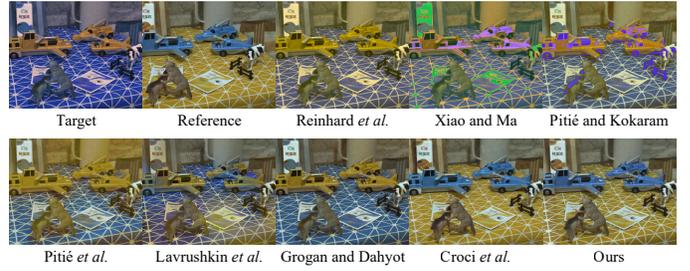}
    \caption{Results of the color transfer from the reference image to the target image on a stereopair from InStereo2K~\cite{bao2020instereo2k}. The hue of the target image was adjusted using the maximum magnitude ($+0.5$). Neural network-based methods (Croci \etal~\cite{croci2021deep} and ours), that were trained on such distortions, have successfully transferred the colors.}
    \label{fig:results}
\end{figure}

\section{Experiments}\label{sec:experiments}

We compared our method with seven others. The comparison included two datasets, four image-quality-assessment methods, and two different training procedures for neural network-based methods.

The first dataset contained undistorted stereopairs from Flickr1024~\cite{wang2019flickr1024} and InStereo2K~\cite{bao2020instereo2k}. To exclude stereopairs with color mismatches, we analyzed them by measuring the mean squared error between two matched views:
\begin{equation}
    \sum_{ij} (T_{ij} - (MR)_{ij})^2 \odot V_{ij} \bigg/ \sum_{ij} V_{ij}.
\end{equation}
After discarding stereopairs with large differences, we obtained 1,035 undistorted stereopairs. We split the dataset into three parts: training (835 stereopairs), validation (100 stereopairs), and test (100 stereopairs). We created an evaluation protocol similar to that used by Croci \etal~\cite{croci2021deep}. Five color operators from TorchVision were employed: brightness, contrast, hue, saturation, and gamma. Each operator was applied to the left view of a stereopair with six magnitude levels, ranging from $-0.5$ to $0.5$. The second dataset, described in Section~\ref{sec:dataset}, contained the distortions we encountered while shooting stereoscopic videos using a beam splitter. We used all 14 videos only for the evaluation.

Our evaluation used four full-reference image-quality-assessment (IQA) methods: peak signal to noise ratio (PSNR), structural similarity (SSIM) index~\cite{wang2004image}, feature similarity for color images (FSIM) index~\cite{zhang2011fsim}, and improved color-image-difference (iCID) metric~\cite{preiss2014color}. To incorporate color information, since PSNR and SSIM work on gray images, we apply them to individual channels of a RGB color space and average channel scores for a final score. A detailed quality analysis of the IQA methods can be found in Niu \etal~\cite{niu2016image} work.

A color distortion model is crucial for neural networks training. Croci \etal~\cite{croci2021deep} used only six color operators, each with six severity levels. This resulted in a network being able to correct 36 distortion types. Therefore, this strategy yielded unsatisfactory results when tested in real-world scenarios. We aim to train a neural network that corrects as many color mismatches as possible. To achieve this, we employed a probabilistic distortion model. We chose six types of distortion operators from TorchVision: brightness, contrast, hue, saturation, gamma, and sharpness. During the training, we applied a random permutation of the above distortions and sampled their parameters from a uniform distribution. This approach enabled the neural network to memorize all distortions within a sampling interval and improved generalization to real-world videos in comparison with Croci's training procedure. To match their number of iterations, we generated 37 training examples per image. Both Croci's and our own training procedures were employed to train neural networks. Table~\ref{tab:comparison} demonstrates that our probabilistic training procedure yielded superior results in out-of-training-domain data.

We trained our network with an Adam optimizer over 100 epochs on the artificial dataset using random patches of size 480x256. Furthermore, random horizontal and vertical flips were applied. When horizontal flips were applied, left and right stereoscopic views were swapped to maintain a stereoscopic property. The initial learning rate was set at $0.0003$ and subsequently decreased to $0.000001$ following a cosine annealing. Training converged in 20 hours on four Nvidia A100 GPUs.

We compared the proposed method with three global methods~\cite{reinhard2001color, xiao2006color, pitie2007linear}, one local method~\cite{pitie2007automated}, two correspondence-based methods~\cite{lavrushkin2018local, grogan2019l2}, and one neural network~\cite{croci2021deep}.  Table~\ref{tab:comparison} and Figure~\ref{fig:results} show results. Artificial dataset's results demonstrate an upper bound of neural networks' ability to correct distortions when a color-distortion model is known. Our method was ranked best by all quality-assessment methods with both training procedures. In contrast, the real-world dataset represents a different domain and serves to benchmark an ability to generalize to unseen data. On this dataset non-global methods, which consider different non-trivial distortion models, performed worse than the simple global methods. We think that this discrepancy is due to the domain shift between the distortion model used in methods' development and the real-world distortion model.

\begin{figure}[tb]
    \centering
    \includegraphics[width=\linewidth]{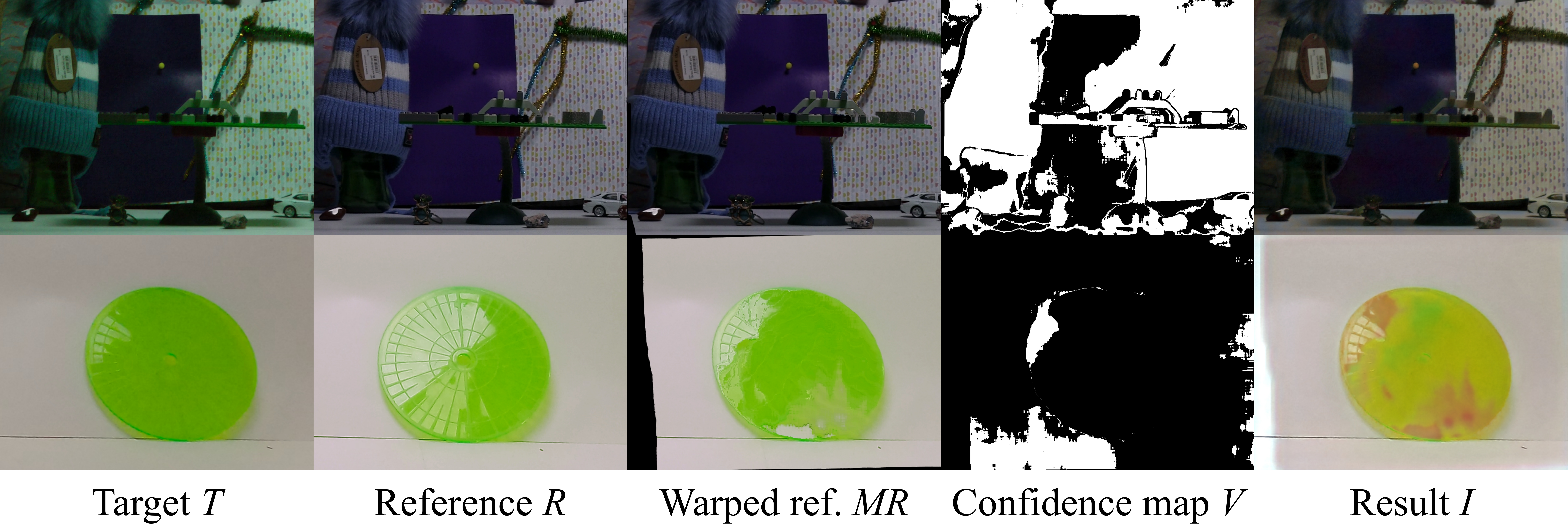}
    \caption{Visualization of our method's failures on a stereopairs from our real-world dataset. The top row contains a scene, that is relatively simple for an optical flow algorithm, and the bottom row contains a difficult scene for it.}
    \label{fig:failures}
\end{figure}

To confirm our claims about domain shift, we have analyzed our deep method piecemeal starting with insights about optical flow algorithm's results. GMFlow---optical flow algorithm that we used---has successfully find an excellent-grade transformation for a simple real-world scene and a fair-grade transformation for a difficult scene with a glare as Figure~\ref{fig:failures} shows. The same applies to the confidence maps, generated by the optical flow algorithm itself. The results of our algorithm, however, should be considered poor, because the color of the result images is not consistent with the color of the reference images even in a high-confidence regions. This confirms that the distortion model, which was considered in the training process, is not sufficient for the cases different from learning domain. The same applies to all methods that consider any color model except global, as Table~\ref{tab:comparison} shows. Future work on this topic definitely should focus on a search for more accurate color-distortion models.

\section{Conclusion}

Few datasets are available for evaluating color-mismatch-correction methods. Most of them are either too small or lack real-world mismatches. To address this, we have created a new dataset of real-world stereoscopic videos. Our proposed method pushed the upper bound of the quality of the color transfer methods on known distortions. We showed that the distortion model, considered in the development and training of the methods, significantly affects the results achieved. We wish our contribution would facilitate the search for accurate color-distortion models and development of methods that generalize better to the unseen color mismatches.

\bibliographystyle{plain}
\bibliography{index}

\end{document}